\pdfoutput=1

\documentclass[11pt]{article}

\usepackage[final]{ACL2023}

\usepackage{times}
\usepackage{latexsym}

\usepackage{graphicx}

\usepackage[T1]{fontenc}

\usepackage[utf8]{inputenc}

\usepackage{microtype}

\usepackage{inconsolata}

\title{DRIVE: Disfluency-Rich Synthetic Dialog Data Generation Framework for Intelligent Vehicle Environments}

\author{
Anshul Chavda\textsuperscript{1} \quad
M Jagadeesh\textsuperscript{2} \quad
Chintalapalli Raja Kullayappa\textsuperscript{2} \quad
B Jayaprakash\textsuperscript{2} \quad \\
\textbf{Medchalimi Sruthi}\textsuperscript{2} \quad
\textbf{Pushpak Bhattacharyya}\textsuperscript{1}
\\[1ex]
\textsuperscript{1}Indian Institute of Technology Bombay \quad  
\textsuperscript{2}Hyundai Motors India Engineering \quad \\
\texttt{\{anshulchavda, pb\}@cse.iitb.ac.in} 
}

\begin{document}
\maketitle

\begin{abstract}
In‐car conversational AI is becoming increasingly critical as autonomous vehicles and smart assistants gain widespread adoption. Yet, existing datasets fail to capture the spontaneous disfluencies—hesitations, false starts, repetitions, and self‐corrections—that characterize real driver–AI dialogs. To address this, we introduce DiscoDrive, a synthetic corpus of 3,500 multi‐turn dialogs across seven automotive domains, generated using a two‐stage, prompt‐driven pipeline that dynamically integrates disfluencies during synthesis. We show that DiscoDrive is effective both as a training resource—enabling DialoGPT-Medium and T5-Base to match or exceed KVRET-trained models on the MultiWOZ 2.2 and Schema-Guided Dialogue (SGD) relevant test sets (BLEU-4 +0.26–0.61; METEOR +2.10; ROUGE-L +3.48; BERTScore F$_1$ +1.35–3.48)—and as a data augmentation resource in low‐resource scenarios, delivering additional gains of up to BLEU-4 +0.38, METEOR +1.95, ROUGE-L +2.87, and BERTScore F$_1$ +4.00 when combined with 10\% of KVRET. Human evaluations further confirm that dialogs sampled from DiscoDrive are rated higher than KVRET’s human‐collected dialogs in naturalness (3.8 vs. 3.6) and coherence (4.1 vs. 4.0), and are perceived as more context‐appropriate than leading post‐hoc methods (e.g., LARD), without compromising clarity. DiscoDrive thus fills a critical gap in existing resources and serves as a versatile corpus for both training and augmenting conversational AI, enabling robust handling of real‐world, disfluent in‐car interactions.
\end{abstract}

\section{Introduction}

Synthetic dialog data generation has become a key area in conversational AI, enabling the creation of large-scale datasets for model training when real-world data is scarce or expensive to collect. Despite significant progress in generating high-quality dialogs for general-purpose assistants like Siri or Alexa, specialized datasets for domains such as automotive applications remain limited. Existing resources, such as KVRET \citep{eric-etal-2017-key}, focus on fluent, task-oriented dialogs across limited domains (e.g., navigation and weather) and do not capture the diversity or spontaneous characteristics needed for more complex, domain-specific scenarios—particularly those involving safety, diagnostics, or urgent requests. This gap is critical, as in-car AI systems must manage unique communication dynamics under high-pressure conditions.

Another key issue is the inherent fluidity of spoken language, which often includes various errors and self-corrections made on the fly. When speakers become aware of an error, they naturally tend to modify, rephrase, or even completely restart their speech. These spontaneous interruptions and self-corrections---commonly referred to as disfluencies \citep{Shriberg1994PreliminariesTA}---are a hallmark of natural conversation. Numerous theoretical studies have investigated the patterns of these disfluencies and how they affect human communication \citep{Sparks1995StructureOS,Shriberg1994PreliminariesTA,Plejert2004FixWhatsNotBroken,Colman2011TheDO,Emrani2019}.

In natural, real-time spoken interactions—such as those in driver-AI systems—disfluencies are inevitable. Studies in spontaneous speech \citep{zayats19_interspeech} and research in naturalistic settings \citep{Seyfeddinipur2006Disfluency} confirm that fillers (e.g., “uh” and “um”), false starts, and self-repairs frequently occur even under high cognitive load. Despite their prevalence, disfluencies remain underrepresented in task-specific dialog datasets. For instance, the widely used Switchboard corpus contains only 5.9\% of tokens annotated as disfluent \citep{charniak-johnson-2001-edit}, limiting exposure to the full range of disfluency phenomena encountered in driver-AI interactions. Moreover, disfluencies disrupt the regular, fluent structure that language models are typically trained on. Research by \citet{FERREIRA2004231} shows that such interruptions can impair syntactic parsing and semantic integration, while \citet{foxtree2001} demonstrate that filler words carry subtle prosodic cues reflecting planning delays—factors that further challenge the robustness of conversational AI.

Consequently, models fine-tuned on fluent, task-oriented datasets may struggle to interpret and respond effectively to the fragmented, disfluent inputs intrinsic to automotive contexts. Addressing this shortfall requires the development of domain-specific synthetic dialog datasets that not only replicate the natural flow of driver-AI interactions but also incorporate a rich variety of disfluencies, ultimately enhancing the naturalness and adaptability of AI systems in real-world automotive environments.

To address the limitations of existing dialog datasets and the unique requirements of human-car AI interactions, we present DiscoDrive (\textbf{DI}sfluency-Enriched \textbf{S}ynthetic \textbf{CO}nversations for \textbf{DRIVE}r-Car AI), a synthetic dialog dataset, with the following contributions:

\begin{enumerate}  
    \item \textbf{DiscoDrive}, a comprehensive dataset of 3,500 dialogs spanning seven domains, each with varying dialog lengths to reflect real-world in-car AI interactions. On average, each dialog contains 8 turns, balancing brief and complex conversations. (Section \ref{sec:methodology})
    \item \textbf{Dynamic disfluency integration} directly during dialog generation via advanced prompt-driven generative modeling, \textbf{eliminating the need for post-hoc disfluency addition.} (Section \ref{sec:methodology}) 
    \item\textbf{Downstream evaluations} on  relevant test subsets of multiWOZ 2.2 \citep{zang-etal-2020-multiwoz} \& SGD \citep{rastogi2020towards}, finding that DialoGPT-Medium and T5-Base fine-tuned solely on DiscoDrive corpus achieve absolute BLEU-4 gains of \textbf{(+0.26 – +0.61)} and BERTScore F$_1$ improvements of \textbf{(+1.35 – +3.48)} over models fine-tuned on KVRET’s training set, while also surpassing them on ROUGE-L and METEOR, \textbf{validating its use as an alternative resource for training in-car conversational AI.} (Section \ref{sec:results}).
    \item In \textbf{low-resource scenarios}, augmenting scarce human data with our DiscoDrive yields substantial uplifts—e.g., BLEU-4 increases by +0.38 and BERTScore F$_1$ by (+3 – +4) over real-only baselines, \textbf{confirming that DiscoDrive is an effective data augmentation resource in low-resource settings} (Section \ref{sec:results}).
    \item \textbf{Extensive human evaluations} reveal that DiscoDrive achieves significantly \textbf{higher ratings in disfluency naturalness and appropriateness compared to a leading post-hoc insertion technique} \citep{passali-etal-2022-lard}, without compromising clarity. Furthermore, it demonstrates \textbf{improvements in naturalness, human-likeness, and lexical diversity compared to the KVRET} \citep{eric-etal-2017-key} baseline (Section \ref{sec:results}). 
\end{enumerate}

\section{Related Work}

\subsection{Synthetic Dialog Data Generation}

Traditional methods have relied on rule-based systems and sequence-to-sequence models, often utilizing manually curated datasets that lack the spontaneous and diverse characteristics of natural conversations. In contrast, recent advances in pre-trained language models like GPT-3 \citep{NEURIPS2020_1457c0d6}, GPT-4o \citep{openai2024gpt4o}, and DialoGPT \citep{zhang-etal-2020-dialogpt} have enabled the generation of more coherent, contextually rich conversations.

Newer frameworks such as PLACES \citep{chen-etal-2023-places} and semi-automated approaches like those presented by \citet{shah-etal-2018-bootstrapping} have further advanced synthetic dialog generation. In addition, recent works have specifically targeted complex conversational phenomena: TOAD \citep{liu-etal-2024-toad} and LUCID \citep{stacey-etal-2024-lucid} generate task-oriented dialogs with diverse response styles and intricate utterance structures, while PRESTO \citep{goel-etal-2023-presto} introduces explicitly labeled, multilingual dialogs that capture realistic conversational dynamics.

\subsection{Disfluency Handling in NLP}

Disfluencies, such as repetitions, false starts, and filler words, are common in spontaneous speech but challenging to model in NLP. Early approaches focused on detection and removal using statistical methods like hidden Markov models and rule-based systems \citep{hough15_interspeech}. With advances in deep learning, models such as bidirectional LSTMs and transformers have achieved better accuracy in detecting disfluencies \citep{zayats16_interspeech}.

While progress has been made in detection, less emphasis has been placed on generating disfluencies for training dialog systems. Recent work, such as \citep{yang-etal-2020-planning}, has explored disfluency generation for data augmentation, while \citep{passali-etal-2022-lard} demonstrated large-scale disfluency generation to enhance model robustness. Additionally, \citep{marie-2023-disfluency} presented methods for generating disfluent dialog to improve the resilience of conversational systems to unstructured user input.

\section{Methodology} \label{sec:methodology}

Our approach to generating synthetic dialogs for driver-AI interactions leverages a two-step process involving conversation scenario generation and dialog simulation. We utilize advanced language models to simulate natural conversational dynamics, including disfluencies, while covering a range of realistic automotive scenarios. The methodology includes turn-based prompting, dynamic disfluency integration, and multi-domain coverage, all aimed at producing a high-quality dataset that reflects the complexities of real-world driver-AI interactions.

\subsection{Two-Step Generation Pipeline}

The synthetic dialog data generation pipeline is structured into two main stages:

\textbf{Step 1: Conversation Scenario Generation.} We generated 500 diverse conversation scenarios for each domain using GPT-4o \citep{openai2024gpt4o}. These scenarios simulate realistic situations for dialogs within the automotive domain, such as "The driver wants to find the shortest route from Mumbai to Pune, avoiding traffic and toll roads." in the Navigation domain. To guide GPT-4o \citep{openai2024gpt4o}, we employed few-shot learning with a curated set of 10–20 human-written examples per domain, ensuring contextual relevance and diversity. The resulting scenarios provide the foundational context for the subsequent dialog simulation stage.

\textbf{Step 2: Dialog Simulation.} Using the Llama-3.1-8B-instruct model \citep{meta2024llama3}, we simulated multi-turn dialogs between the driver and the car AI. The simulation alternates between the two roles, with prompts tailored to each speaker’s conversational style—informal and disfluent for the driver, concise and task-focused for the car AI. Conversation history was limited to the last six exchanges to maintain coherence while balancing computational efficiency. This process ensured that the generated dialogs were contextually appropriate and aligned with the scenario.

\subsection{Turn-Based Prompting for Dialog Simulation}

The dialog simulation process utilized a turn-based framework, alternating between prompts for the driver and the car AI to create realistic and contextually coherent exchanges.

\textbf{Driver Prompts:} Driver prompts were crafted to emulate the characteristics of natural communication in driving scenarios. These prompts simulated scenarios where the driver might be multitasking or distracted, incorporating informal language and natural disfluencies such as hesitations, repetitions, and self-corrections. This approach ensured that the generated dialogs reflected the spontaneity and variability of real-world interactions.

\textbf{Car AI Prompts:} The car AI prompts were designed to produce responses that are concise, informative, and aligned with the driver’s requests. The tone was deliberately conversational and friendly, reflecting the role of an in-car assistant, while prioritizing clarity and efficiency in addressing the driver's needs.

At each turn, the model generated text based on the evolving conversation history, allowing it to leverage prior exchanges to maintain coherence and contextual relevance. Dynamic adjustments were made to the prompts during the dialog generation process, tailoring them to the specific stage of the conversation. For regular turns, prompts ensured a natural flow of exchanges, while for concluding turns, prompts guided the model to wrap up the dialog concisely and contextually. All prompt templates used to generate DiscoDrive are provided in Appendix~\ref{sec:appendix-prompts}.

\begin{table*}[ht!]
\centering
\resizebox{\textwidth}{!}{%
\begin{tabular}{c l p{6cm} p{6cm}}  
\hline
\textbf{No.} & \textbf{Type} & \textbf{Description} & \textbf{Example} \\ 
\hline
1 & Repetitions & Repeating a word or phrase to simulate hesitation or emphasize a point. & "I think, I think we should take the next exit." \\ 
2 & False Starts & Beginning a thought but switching direction mid-sentence, reflecting spontaneous speech corrections. & "We could—actually, let’s try the other route." \\ 
3 & Filler Words & Incorporating words like "um" or "uh" to represent pauses in thinking or uncertainty. & "Can you, um, check the tire pressure?" \\ 
4 & Pauses & Using ellipses ("...") to indicate brief pauses or hesitation in speech. & "I think we’ll be there... um, soon." \\ 
5 & Corrections & Revising a previously stated idea or instruction to reflect common human speech adjustments. & "Turn left—no, wait, I mean right." \\ 
\hline
\end{tabular}%
}
\caption{Categories of Disfluencies Used in Dialog Generation: Definitions and Examples.}
\label{tab:disfluency_types}
\end{table*}

\subsection{Dynamic Disfluency Integration}

We dynamically integrated disfluencies during the generation process via prompt-driven generative modeling. This approach is motivated by extensive linguistic research demonstrating that natural speech is inherently disfluent.

For instance, repetitions frequently occur when speakers hesitate or emphasize a point, reflecting underlying cognitive load or uncertainty \citep{Shriberg1994PreliminariesTA}. Similarly, false starts and subsequent corrections are common in spontaneous conversation, as speakers often self-edit mid-utterance to clarify their intended message \citep{Sparks1995StructureOS, Plejert2004FixWhatsNotBroken, Colman2011TheDO}. Filler words like “um” and “uh” are pervasive in everyday speech and serve as markers of planning delays or uncertainty, providing listeners with subtle prosodic cues \citep{foxtree2001, FERREIRA2004231}. Additionally, natural pauses—represented in transcripts by ellipses—occur as speakers momentarily pause to organize their thoughts.

The types and examples of these disfluencies are detailed in Table~\ref{tab:disfluency_types}. By incorporating these empirically validated disfluency types, our synthetic dialogs more accurately capture the spontaneous, variable nature of real-world driver–AI interactions, thereby enhancing the robustness and adaptability of conversational AI systems.

\subsection{Domains and Scenarios}

To ensure comprehensive coverage of automotive-related interactions, our dataset spans seven distinct domains:

\begin{enumerate} 
    \item \textbf{Navigation:} Involving requests for directions, traffic updates, and finding routes. 
    \item \textbf{Car Maintenance and Diagnostics:} Addressing vehicle status checks, troubleshooting, and scheduling services. \item \textbf{Safety and Emergency Assistance:} Focusing on alerts, emergency handling, and safety-related inquiries. 
    \item \textbf{Entertainment:} Dialogs about media controls, music, radio, and other entertainment options. 
    \item \textbf{Local and On-Route Attractions and Activities:} Providing information on nearby points of interest, restaurants, and events. 
    \item \textbf{Car Functions:} Interactions about vehicle controls, such as adjusting air conditioning or seat settings. \item \textbf{Weather:} Queries related to current weather conditions, forecasts, and weather-related alerts. 
\end{enumerate}

For each domain, we generated 500 unique conversation scenarios using GPT-4o \citep{openai2024gpt4o}, resulting in a total of 3,500 dialogs. The dialogs vary in length (6, 8, 10, 12, 14 turns) to reflect the range of real-world driver-AI interactions, from brief task-oriented exchanges to more complex, context-rich discussions. This range balances realism and diversity, ensuring the dataset captures the practical nuances of automotive conversations while supporting robust model training.

\section{Experimental Setup}

\subsection{Compute Infrastructure}
We ran generation on NVIDIA A40 and A100 GPUs (48 GB and 80 GB VRAM) using bfloat16 precision.  Scenario and dialog synthesis consumed 40 GPU-hours; downstream fine-tuning (DialoGPT-Medium \citep{zhang-etal-2020-dialogpt}, T5-Base \citep{raffel2019exploring}) and inference together used 30 GPU-hours.

\subsection{Software Stack}
Our pipeline is implemented in Python with HuggingFace Transformers v4.x.  We use GPT-4o \citep{openai2024gpt4o} for scenario generation and LLaMA-3.1-8B-Instruct \citep{meta2024llama3} for simulation.  Downstream models (DialoGPT-Medium \citep{zhang-etal-2020-dialogpt}, T5-Base \citep{raffel2019exploring}) are fine-tuned via HuggingFace’s Seq2SeqTrainer and Trainer APIs, with dynamic padding, gradient accumulation (batch size 16), and AdamW (lr = 5e-5 for DialoGPT-Medium \citep{zhang-etal-2020-dialogpt}, 3e-5 for T5-Base \citep{raffel2019exploring}) to ensure efficient multi-GPU training.

\section{Results} \label{sec:results}

This section presents the results and key observations from our experiments and evaluations.

\subsection{Automatic Evaluation}

To evaluate the lexical richness of our DiscoDrive dataset compared to KVRET dataset \citep{eric-etal-2017-key}, we employed N-distinct \citep{li-etal-2016-diversity}, a widely-used metric for assessing lexical diversity. 
The results of the N-Distinct evaluation are presented in Table~\ref{tab:ndistinct}.  

\begin{table}[ht!]
\centering
\small
\resizebox{\columnwidth}{!}{%
\begin{tabular}{|p{1.5cm}|c|c|}
\hline
\textbf{N-Gram} & \textbf{KVRET} & \textbf{DiscoDrive Dataset} \\
\hline
1-gram & 0.0109 & 0.0124 \\
\hline
2-gram & 0.1040 & 0.1234 \\
\hline
3-gram & 0.2902 & 0.3428 \\
\hline
4-gram & 0.4808 & 0.5425 \\
\hline
\end{tabular}%
}
\caption{N-Distinct scores for lexical diversity in KVRET \citep{eric-etal-2017-key} and DiscoDrive.}
\label{tab:ndistinct}
\end{table}

\textbf{Observations:} DiscoDrive achieves consistently higher N-Distinct scores across all n-gram levels compared to KVRET \citep{eric-etal-2017-key}, demonstrating higher lexical diversity. 

However, it is important to note that lexical diversity alone does not capture other essential aspects of dialog quality, such as naturalness and coherence, which are addressed through human evaluations in this work.

\subsection{Human Evaluation}

To evaluate the quality of DiscoDrive, two human evaluators performed individual dialog assessments, pairwise comparisons, and a targeted evaluation of dynamic versus post-hoc disfluency integration. For these evaluations, we used stratified subsets of 140 dialogues each from the DiscoDrive and KVRET datasets \citep{eric-etal-2017-key}, ensuring equal representation across domains.
A detailed description of the human evaluation setup can be found in Section \ref{sec:appendix-human-eval}.

\subsubsection{Intrinsic Evaluation Results}
Human evaluators rated the dialogs based on various criteria using a 5-point Likert scale. The results are summarized in Table \ref{tab:intrinsic-eval}.

\begin{table}[ht!]
\small
\centering
\resizebox{\columnwidth}{!}{%
\begin{tabular}{|p{2cm}|c|c|}
\hline
\textbf{Metric} & \textbf{KVRET} & \textbf{DiscoDrive} \\
\hline
Naturalness & 3.6 ($\pm$0.18) & 3.8 ($\pm$0.18)\\
\hline
Coherence & 4.0 ($\pm$0.17) & 4.1 ($\pm$0.16)\\
\hline
Engagement & 4.0 ($\pm$0.17) & 3.8 ($\pm$0.17)\\
\hline
Consistency & 4.3 ($\pm$0.15) & 4.2 ($\pm$0.16)\\
\hline
On-topic & 4.9 ($\pm$0.04) & 4.7 ($\pm$0.10)\\
\hline
\end{tabular}%
}
\caption{Intrinsic evaluation results based on a 5-point Likert scale. The values represent the average scores from two human evaluators and the confidence intervals.}
\label{tab:intrinsic-eval}
\end{table}

\textbf{Observations:}
DiscoDrive dataset outperforms KVRET \citep{eric-etal-2017-key} in naturalness (3.8 vs. 3.6) and coherence (4.1 vs. 4.0), highlighting the positive impact of dynamically integrated disfluencies on dialog realism. However, the slight drop in engagement (3.8 vs. 4.0) and on-topic relevance (4.7 vs. 4.9) reflects the inherent variability introduced by disfluencies, which can occasionally detract from task focus. Despite these trade-offs, the synthetic dataset maintains strong consistency (4.2 vs. 4.3), demonstrating its suitability for training models that balance conversational realism and task-oriented performance.

\subsubsection{Comparative Evaluation Results}
\begin{figure}[ht!]
    \centering
    \includegraphics[width=\columnwidth]{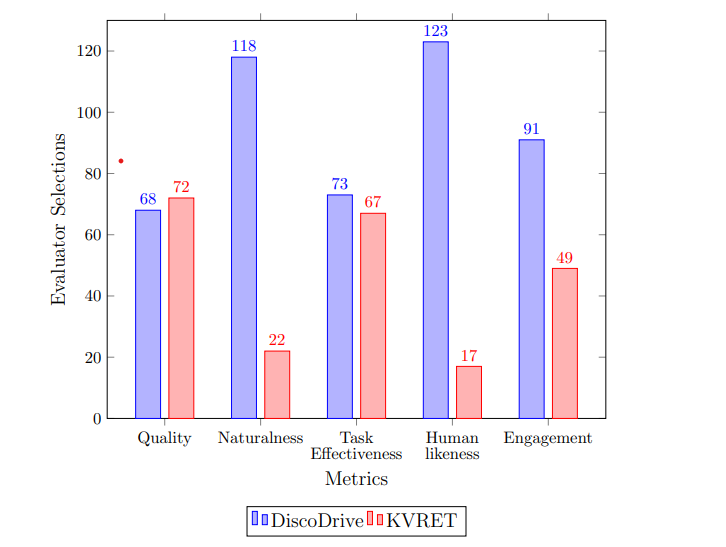}
    \caption{Comparative evaluation results for KVRET and DiscoDrive. The values represent the average scores from two human evaluators.}
    \label{fig:comparative-eval}
\end{figure}

In pairwise comparisons, evaluators selected which dialog was better across various criteria. The results are visualized in Figure \ref{fig:comparative-eval}.

\textbf{Observations:} The DiscoDrive dataset outperformed KVRET \citep{eric-etal-2017-key} in naturalness (118 vs. 22), human-likeness (123 vs. 17), and engagement (91 vs. 49), demonstrating the success of dynamically integrated disfluencies in producing more human-like, spontaneous dialogs.

While KVRET showed a slight edge in overall quality (72 vs. 68), likely due to its structured design, the synthetic dataset remained competitive in task effectiveness (73 vs. 67), balancing realism and task relevance effectively.

These results highlight the trade-off between conversational variability and structure but affirm the DiscoDrive’s strength in delivering realistic and engaging dialogs.

\subsubsection{Evaluation of Dynamic vs. Post-Hoc Disfluency Integration}
To further evaluate the impact of the disfluency integration approach, we applied the LARD method \citep{passali-etal-2022-lard} for post-hoc insertion of disfluencies to the driver utterances in the KVRET \citep{eric-etal-2017-key} dialogs. Using the same subset of 140 dialogs employed in our other human evaluations, two human evaluators rated the dialogs on a 5-point Likert scale for various metrics. The results are summarized in Table~\ref{tab:disfluency_comparison}.

\begin{table}[ht!]
\small
\centering
\resizebox{\columnwidth}{!}{%
\begin{tabular}{|l|c|c|}
\hline
\textbf{Metric} & \textbf{KVRET} & \textbf{DiscoDrive} \\ 
\hline
Naturalness & 3.6 ($\pm$0.17) & 4.2 ($\pm$0.16)\\ 
\hline
Appropriateness & 3.4 ($\pm$0.18) & 4.3 ($\pm$0.15)\\ 
\hline
Clarity & 4.2 ($\pm$0.16) & 4.0 ($\pm$0.16)\\ 
\hline
\end{tabular}
}
\caption{Evaluation results for KVRET dialogs with post-hoc disfluency insertion using LARD \citep{passali-etal-2022-lard} and DiscoDrive dialogs with dynamic disfluency integration. The values represent the average scores from two human evaluators and the confidence intervals.}
\label{tab:disfluency_comparison}
\end{table}

\textbf{Observations:}
The results indicate that dialogs generated with dynamic disfluency integration are perceived as significantly more natural and contextually appropriate than those modified using the LARD method \citep{passali-etal-2022-lard}. While clarity remains comparable between the two methods, the marked improvements in naturalness (4.2 vs. 3.6) and appropriateness (4.3 vs. 3.4) validate our approach for enhancing the realism of driver-AI interactions. Examples of post-hoc disfluency insertions using LARD method \citep{passali-etal-2022-lard} are provided in Appendix~\ref{sec:post-hoc}.

\subsection{Experimental Evaluation of Full-Data Fine-Tuning for Car AI Response Generation}
We assess the utility of our DiscoDrive dataset as a complete training resource by fine-tuning two standard dialogue models—DialoGPT-Medium \citep{zhang-etal-2020-dialogpt} and T5-Base \citep{raffel2019exploring}—under three conditions: (i) zero-shot (no fine-tuning), (ii) fine-tuned on the full KVRET training split (2,424 dialogs), and (iii) fine-tuned on our full DiscoDrive corpus (3,500 dialogs). We evaluate on dialog subsets (220 dialogs each) curated from multiWOZ 2.2 \citep{zang-etal-2020-multiwoz} and SGD \citep{rastogi2020towards} by filtering for services aligned with in-car assistant functionality, including domains such as navigation, weather updates, hotel reservations, attraction search, and restaurant inquiries. Evaluation metrics include BLEU-1–4, ROUGE-L, METEOR, and BERTScore F$_1$.

\begin{table*}[t]
  \centering
  \small
  \begin{tabular}{lccccccc}
    \hline
    {\bf Model \& Configuration} 
      & \bf BLEU-1 & \bf BLEU-2 & \bf BLEU-3 & \bf BLEU-4 
      & \bf ROUGE-L & \bf METEOR & \bf BERTScore F$_1$ \\
    \hline
    \multicolumn{8}{l}{\emph{DialoGPT-Medium}} \\
    \hline
    Zero–Shot               & 01.10 & 00.14 & 00.05 & 00.01 &  3.64 &  2.24 & 74.82 \\
    KVRET Fine-Tuned        & 05.55 &  0.51 &  0.14 &  0.03 &  9.41 &  6.46 & 82.71 \\
    Synthetic Fine-Tuned    & \textbf{14.10} & \textbf{3.18} & \textbf{1.39} & \textbf{0.64} 
                            & \textbf{16.11} & \textbf{20.51} & \textbf{85.20} \\
    \hline
    \multicolumn{8}{l}{\emph{T5-Base}} \\
    \hline
    Zero–Shot               & 12.34 &  2.10 &  0.95 &  0.42 & 10.20 & 12.80 & 78.30 \\
    KVRET Fine-Tuned        & \textbf{18.30} &  2.90 &  1.23 &  0.61 & 13.05 &  9.28 & 84.13 \\
    Synthetic Fine-Tuned    & 16.28 & \textbf{3.85} & \textbf{1.64} & \textbf{0.77} 
                            & \textbf{16.19} & \textbf{22.78} & \textbf{85.48} \\
    \hline
  \end{tabular}
  \caption{Full-data fine-tuning results on the multiWOZ 2.2 in-car subset (beam=5). DiscoDrive-only fine-tuning outperforms KVRET on nearly all metrics.}
  \label{tab:full-mwoz}
\end{table*}

\begin{table*}[t]
  \centering
  \small
  \begin{tabular}{lccccccc}
    \hline
    {\bf Model \& Configuration}
    & \bf BLEU-1 & \bf BLEU-2 & \bf BLEU-3 & \bf BLEU-4 
      & \bf ROUGE-L & \bf METEOR & \bf BERTScore F$_1$ \\
    \hline
    \multicolumn{8}{l}{\emph{DialoGPT-Medium}} \\
    \hline
    Zero–Shot               & 00.82 & 00.09 & 00.02 & 00.00 &  2.90 &  1.94 & 75.08 \\
    Fine-Tuned on KVRET        & 01.91 &  0.18 &  0.06 &  0.02 &  7.03 &  5.21 & 81.23 \\
    Fine-Tuned on DiscoDrive    & \textbf{9.98} & \textbf{1.18} & \textbf{0.68} & \textbf{0.28} 
                            & \textbf{12.35} & \textbf{17.57} & \textbf{84.71} \\
    \hline
    \multicolumn{8}{l}{\emph{T5-Base}} \\
    \hline
    Zero–Shot               & 08.28 & 01.39 & 00.56 & 00.25 &  8.60 & 14.16 & 82.98 \\
    Fine-Tuned on KVRET        & \textbf{15.39} &  2.86 &  1.31 &  0.67 & 10.51 &  8.03 & 83.83 \\
    Fine-Tuned on DiscoDrive    & 14.47 & \textbf{3.91} & \textbf{1.86} & \textbf{0.90} 
                            & \textbf{15.28} & \textbf{22.54} & \textbf{85.55} \\
    \hline
  \end{tabular}
  \caption{Full-data fine-tuning results on the SGD in-car subset (beam=5). DiscoDrive-only fine-tuning outperforms KVRET on nearly all metrics.}
  \label{tab:full-sgd}
\end{table*}

\paragraph{multiWOZ 2.2 Results:}  
Table \ref{tab:full-mwoz} shows that models trained on the DiscoDrive corpus consistently outperform their KVRET-trained counterparts across BLEU, ROUGE, METEOR, and BERTScore. For DialoGPT \citep{zhang-etal-2020-dialogpt}, fine-tuning on DiscoDrive yields a 21× improvement in BLEU-4 compared to KVRET \citep{eric-etal-2017-key} (0.64 vs. 0.03), and a +2.5 point increase in BERTScore F$_1$ (85.20 vs. 82.71). T5-Base \citep{raffel2019exploring} shows a similar trend: although KVRET \citep{eric-etal-2017-key} leads slightly on BLEU-1, the DiscoDrive-trained model performs better on higher-order BLEU (BLEU-2–4), METEOR (↑+13.5 points), ROUGE-L (↑+3.1), and BERTScore F$_1$ (↑+1.35).

\paragraph{SGD Results:}  
Table \ref{tab:full-sgd} reports that the performance gap is even more pronounced. DialoGPT \citep{zhang-etal-2020-dialogpt} fine-tuned on DiscoDrive achieves a BLEU-4 of 0.28, substantially outperforming the KVRET-trained model (0.02), alongside a +3.5 point gain in BERTScore F$_1$. For T5-Base \citep{raffel2019exploring}, DiscoDrive fine-tuning yields the strongest results across all metrics except BLEU-1, including +0.59 BLEU-4 and +1.72 BERTScore F$_1$ over the KVRET-trained version.

\paragraph{Analysis:}  
Across both multiWOZ 2.2 \citep{zang-etal-2020-multiwoz} and SGD subsets \citep{rastogi2020towards}, and for both DialoGPT-Medium \citep{zhang-etal-2020-dialogpt} and T5-Base \citep{raffel2019exploring}, fine-tuning on the DiscoDrive corpus produces generation quality that consistently surpasses that achieved using the human-curated KVRET dataset \citep{eric-etal-2017-key} (Tables \ref{tab:full-mwoz}–\ref{tab:full-sgd}). These results confirm the robustness and versatility of our synthetic dialogs, demonstrating DiscoDrive as a viable, scalable alternative to real-world annotations for comprehensive, full-scale training and rapid domain adaptation in conversational AI.  

\subsection{Experimental Evaluation of Low-Resource Data Augmentation for Car AI Response Generation}

To evaluate the utility of our DiscoDrive dataset in low-resource scenarios, we fine-tune DialoGPT-Medium \citep{zhang-etal-2020-dialogpt} and T5-Base \citep{raffel2019exploring} under two configurations: (i) using only 10\% of KVRET’s training split (242 dialogs), and (ii) augmenting the same 10\% KVRET subset with the full DiscoDrive corpus (3,500 dialogs). We evaluate on the same dialog subsets (220 dialogs each) curated from multiWOZ 2.2 \citep{zang-etal-2020-multiwoz} and SGD \citep{rastogi2020towards} as used in the full-data experiments. Evaluation metrics include BLEU-1–4, ROUGE-L, METEOR, and BERTScore F$_1$. Table \ref{tab:lowres-finetune-mwoz} \& \ref{tab:lowres-finetune-sgd} presents the low-resource results. 

\paragraph{multiWOZ 2.2 Results:}  
As reported in Table \ref{tab:lowres-finetune-mwoz}, DiscoDrive augmentation substantially improves generation quality across both models. For DialoGPT-Medium \citep{zhang-etal-2020-dialogpt}, adding synthetic dialogs increases BLEU-4 from 0.05 to 0.58 and boosts BERTScore F$_1$ from 81.96 to 85.40—narrowing the gap toward full-data performance. For T5-Base \citep{raffel2019exploring}, we observe similar gains in higher-order BLEU, METEOR (↑+14.6), ROUGE-L (↑+7.6), and BERTScore F$_1$ (↑+2.4). These gains validate the complementarity of synthetic disfluency-rich dialogs in enhancing low-data generalization.

\paragraph{SGD Results:}  
Table \ref{tab:lowres-finetune-sgd} reveals analogous gains on the SGD in-car subset: DialoGPT-Medium \citep{zhang-etal-2020-dialogpt} achieves a BLEU-4 increase from 0.00 to 0.88 and a +4.9-point rise in BERTScore F$_1$, while T5-Base \citep{raffel2019exploring} sees its BLEU-4 jump from 1.06 to 1.26 and records METEOR and ROUGE-L improvements exceeding +10 points each.

\begin{table*}[ht!]
\centering
\small
\begin{tabular}{lccccccc}
\hline
{\bf Model \& Configuration} & \bf BLEU-1 & \bf BLEU-2 & \bf BLEU-3 & \bf BLEU-4 & \bf ROUGE-L & \bf METEOR & \bf BERTScore F$_1$ \\
\hline
\multicolumn{8}{l}{\emph{DialoGPT-Medium}} \\
\hline
10\% KVRET Only           & 03.81 & 0.43 & 0.15 & 0.05 &  8.57 & 6.85 & 81.96 \\
10\% KVRET + Synthetic    & \textbf{18.47} & \textbf{3.99} & \textbf{1.52} & \textbf{0.58} & \textbf{16.21} & \textbf{14.96} & \textbf{85.40} \\
\hline
\multicolumn{8}{l}{\emph{T5-Base}} \\
\hline
10\% KVRET Only           & \textbf{23.42} & 3.29 & 1.81 & 1.97 & 10.3 & 6.16 & 83.45 \\
10\% KVRET + Synthetic    & 19.85 & \textbf{5.10} & \textbf{2.25} & \textbf{1.04} & \textbf{17.92} & \textbf{20.76} & \textbf{85.84} \\
\hline
\end{tabular}
\caption{Low-resource fine-tuning results on the multiWOZ 2.2 in-car subset (beam=5). Synthetic augmentation improves BLEU-4, ROUGE-L, METEOR, and BERTScore.}
\label{tab:lowres-finetune-mwoz}
\end{table*}

\begin{table*}[ht!]
\centering
\small
\begin{tabular}{lccccccc}
\hline
{\bf Model \& Configuration} & \bf BLEU-1 & \bf BLEU-2 & \bf BLEU-3 & \bf BLEU-4 & \bf ROUGE-L & \bf METEOR & \bf BERTScore F$_1$ \\
\hline
\multicolumn{8}{l}{\emph{DialoGPT-Medium}} \\
\hline
10\% KVRET Only           & 1.38 & 0.10 & 0.02 & 0.00 &  5.83 & 4.44 & 80.44 \\
10\% KVRET + Synthetic    & \textbf{17.57} & \textbf{3.86} & \textbf{1.84} & \textbf{0.88} & \textbf{14.02} & \textbf{13.46} & \textbf{85.37} \\
\hline
\multicolumn{8}{l}{\emph{T5-Base}} \\
\hline
10\% KVRET Only           & 17.32 & 1.58 & 1.06 & 1.06 & 7.80 & 5.46 & 83.36 \\
10\% KVRET + Synthetic    & \textbf{17.76} & \textbf{5.12} & \textbf{2.53} & \textbf{1.26} & \textbf{16.60} & \textbf{20.40} & \textbf{85.83} \\
\hline
\end{tabular}
\caption{Low-resource fine-tuning results on the SGD in-car subset (beam=5). Synthetic augmentation improves BLEU-4, ROUGE-L, METEOR, and BERTScore.}
\label{tab:lowres-finetune-sgd}
\end{table*}

\paragraph{Analysis:}  
Across both evaluation sets, multiWOZ 2.2 \citep{zang-etal-2020-multiwoz} and SGD \citep{rastogi2020towards}—and for both DialoGPT-Medium \citep{zhang-etal-2020-dialogpt} and T5-Base \citep{raffel2019exploring}, augmenting a small fraction of real dialogs with synthetic data leads to substantial quality gains (Tables \ref{tab:lowres-finetune-mwoz}–\ref{tab:lowres-finetune-sgd}). This demonstrates that our synthetic corpus is not only effective in full-resource conditions but also provides strong augmentation value in low-resource settings where human data is limited or expensive to collect.

\section{Conclusion and Future Work}

We introduced DiscoDrive, a synthetic dialog dataset specifically designed for driver-AI interactions that robustly captures the spontaneous and disfluent nature of real-world speech. Our approach integrates disfluencies—such as hesitations, repetitions, and self-corrections—directly during the generation process via prompt-driven generative modeling, thereby obviating the need for less coherent post-hoc modifications. Spanning \textbf{seven diverse automotive domains}, including navigation, diagnostics, and safety, the dataset offers comprehensive coverage of realistic in-car scenarios. Evaluation across multiple axes validates validates its effectiveness as a resource for training conversational AI systems capable of handling spontaneous, disfluent inputs in complex real-world environments. DiscoDrive outperforms the human-collected KVRET \citep{eric-etal-2017-key} corpus in terms of naturalness (↑0.2 Likert), coherence, and lexical diversity (Distinct-4 ↑0.06), and it significantly exceeds LARD-style \citep{passali-etal-2022-lard} post-hoc insertion techniques in disfluency naturalness (+0.6 Likert) and contextual appropriateness (+0.9 Likert). In downstream tasks, models fine-tuned solely on DiscoDrive match or outperform KVRET-trained models, achieving absolute BLEU-4 gains of +0.3 to +0.6 and BERTScore F$_1$ improvements of +1.4 to +3.5 across the multiWOZ 2.2 \citep{zang-etal-2020-multiwoz} and SGD \citep{rastogi2020towards} in-car subsets. Under low-resource settings (10\% of KVRET \citep{eric-etal-2017-key}), adding DiscoDrive to the limited real data yields substantial gains, with BLEU-4 increases of +0.38 and BERTScore F$_1$ gains of +3–4 points over real-only baselines.

While DiscoDrive establishes the value of synthetic disfluency-rich dialogs for training and augmenting in-car conversational agents, several avenues remain for extension. First, future work can explore \textit{adaptive disfluency control}, where the type and frequency of disfluencies are conditioned on dialog context, user state, or domain complexity. Second, we aim to develop \textit{disfluency-aware evaluation metrics} that jointly assess both task success and naturalness in noisy conversational settings—extending beyond surface-form metrics like BLEU or ROUGE. Third, extending DiscoDrive to new domains (e.g., healthcare, finance, customer service) using prompt-only domain adaptation would test the generality of our pipeline and help validate cross-domain robustness. Finally, integrating the dataset into real-world in-car assistants for user-facing evaluations will offer critical feedback on system usability, dialog resilience, and perceived naturalness in production settings.

\section*{Limitations}

While our synthetic dialog dataset demonstrates promising results in advancing the naturalness of in-car conversational AI, several limitations remain that warrant further investigation.

\begin{enumerate}
    \item \textbf{Task-Oriented Clarity:} Although disfluencies contribute to more human-like interactions, their inclusion introduces potential challenges in critical task-oriented scenarios, such as navigation and safety-related tasks. Disfluencies can create ambiguity or slow down the dialog, potentially hindering clear communication when urgency and precision are required. Managing the balance between realism and clarity remains a key challenge for future work.
    
    \item \textbf{Language Constraints:} The dataset’s focus on English-language interactions limits its applicability to non-English speaking markets. Given the global use of automotive technologies, developing multilingual disfluency-enriched datasets is essential to ensure broader applicability. Models trained exclusively on English data may struggle to generalize to languages with different syntactic or prosodic structures.
    
    \item \textbf{Synthetic Data Limitations:} While we dynamically generate disfluencies, real-world data may present more complex and less predictable patterns of speech. Further evaluation in real-world driving scenarios is necessary to determine how well the synthetic dialogs simulate authentic human-driver interactions.
    
    \item \textbf{Resource-Intensive Models:} The use of large-scale language models, such as GPT-4o \citep{openai2024gpt4o} and Llama-3.1-8B-instruct \citep{meta2024llama3}, presents a significant computational cost, which may limit their real-time applicability in low-resource environments. Developing more efficient strategies or adopting smaller models could address these challenges for deployment in practical applications.

\end{enumerate}

These limitations underscore critical areas for further research and improvements to enhance the realism, applicability, and efficiency of the dataset in real-world automotive systems.

\section*{Ethical Statement}

\textbf{Human Evaluation and Crowdsourcing.} Our human evaluation involved two annotators who assessed both the DiscoDrive and KVRET \citep{eric-etal-2017-key} dataset dialogs. While this study did not employ crowdsourcing platforms, it followed ethical guidelines by ensuring voluntary participation of evaluators. No personal data or sensitive information was collected during this evaluation process. Evaluators were fully informed of their roles and the nature of the tasks, ensuring transparency throughout the assessment.

\textbf{Bias and Language Model Considerations.} The synthetic dialogs were generated using large pre-trained language models, such as Llama-3.1-8B-instruct \citep{meta2024llama3}. These models were trained on vast corpora, which may contain biases or undesirable content originating from the data sources. We recognize that language models can perpetuate or even amplify societal biases present in the training data. Although our work aims to generate task-specific, human-like dialogs, care must be taken to evaluate and mitigate potential bias or inappropriate content in future applications of this dataset. Incorporating bias detection and fairness in the training pipeline is critical for ensuring ethical AI development.

\textbf{Safety in Automotive Applications.} Since the dialogs are tailored for use in automotive environments, safety is a major concern. In-car AI systems must be designed to handle user inputs effectively, particularly in stressful or high-stakes situations. Misinterpretations or inappropriate responses could lead to serious consequences. Our dataset simulates natural human-like interactions, but additional safety checks should be incorporated when deploying these systems to ensure they meet high standards of accuracy and reliability in real-world applications.

\textbf{Scientific Artifacts.} All scientific artifacts were used in accordance with their intended licenses. The KVRET \citep{eric-etal-2017-key} dataset is publicly available and can be accessed at \url{http://nlp.stanford.edu/projects/kvret/kvret_dataset_public.zip}. MultiWOZ 2.2 \citep{zang-etal-2020-multiwoz} and the Schema-Guided Dialogue (SGD) dataset \citep{rastogi2020towards} are publicly released benchmarks widely adopted for multi-domain task-oriented dialog modeling. LLaMA-3.1-8B-Instruct is an open-source language model released by Meta AI for non-commercial research purposes \citep{meta2024llama3}. GPT-4o was accessed via OpenAI’s free-tier research API in compliance with usage terms \citep{openai2024gpt4o}. We also fine-tuned and evaluated DialoGPT-Medium \citep{zhang-etal-2020-dialogpt} and T5-Base \citep{raffel2019exploring}, both of which are publicly available under non-commercial licenses through the HuggingFace Hub. We used HuggingFace Transformers \citep{wolf-etal-2020-transformers} and PyTorch \citep{NEURIPS2019_bdbca288} as the primary libraries for implementation, model training, and data preprocessing. All artifacts and models utilized in this study were English-language based. In line with ethical standards, all datasets were processed in compliance with their terms of use, with no re-identification of participants and full respect for data privacy.

\textbf{Data Privacy and Security.} While the dialogs in our dataset are synthetically generated and do not involve real user data, future deployments of conversational AI systems in cars must ensure compliance with data privacy laws and secure data handling practices.

\nocite{zhang-etal-2020-dialogpt, eric-etal-2017-key, hough15_interspeech, yang-etal-2020-planning, passali-etal-2022-lard, braun-etal-2017-evaluating, abdullin-etal-2023-synthetic, marie-2023-disfluency, zayats16_interspeech, chen-etal-2023-places, li-etal-2016-diversity, meta2024llama3, openai2024gpt4o, NEURIPS2020_1457c0d6, NEURIPS2019_bdbca288}

\bibliographystyle{acl_natbib}
\bibliography{anthology, custom}

\begin{thebibliography}{33}
\expandafter\ifx\csname natexlab\endcsname\relax\def\natexlab#1{#1}\fi

\bibitem[{Abdullin et~al.(2023)Abdullin, Molla, Ofoghi, Yearwood, and Li}]{abdullin-etal-2023-synthetic}
Yelaman Abdullin, Diego Molla, Bahadorreza Ofoghi, John Yearwood, and Qingyang Li. 2023.
\newblock \href {https://aclanthology.org/2023.gem-1.16} {Synthetic dialogue dataset generation using {LLM} agents}.
\newblock In \emph{Proceedings of the Third Workshop on Natural Language Generation, Evaluation, and Metrics (GEM)}, pages 181--191, Singapore. Association for Computational Linguistics.

\bibitem[{Braun et~al.(2017)Braun, Hernandez~Mendez, Matthes, and Langen}]{braun-etal-2017-evaluating}
Daniel Braun, Adrian Hernandez~Mendez, Florian Matthes, and Manfred Langen. 2017.
\newblock \href {https://doi.org/10.18653/v1/W17-5522} {Evaluating natural language understanding services for conversational question answering systems}.
\newblock In \emph{Proceedings of the 18th Annual {SIG}dial Meeting on Discourse and Dialogue}, pages 174--185, Saarbr{\"u}cken, Germany. Association for Computational Linguistics.

\bibitem[{Brown et~al.(2020)Brown, Mann, Ryder, Subbiah, Kaplan, Dhariwal, Neelakantan, Shyam, Sastry, Askell, Agarwal, Herbert-Voss, Krueger, Henighan, Child, Ramesh, Ziegler, Wu, Winter, Hesse, Chen, Sigler, Litwin, Gray, Chess, Clark, Berner, McCandlish, Radford, Sutskever, and Amodei}]{NEURIPS2020_1457c0d6}
Tom Brown, Benjamin Mann, Nick Ryder, Melanie Subbiah, Jared~D Kaplan, Prafulla Dhariwal, Arvind Neelakantan, Pranav Shyam, Girish Sastry, Amanda Askell, Sandhini Agarwal, Ariel Herbert-Voss, Gretchen Krueger, Tom Henighan, Rewon Child, Aditya Ramesh, Daniel Ziegler, Jeffrey Wu, Clemens Winter, Chris Hesse, Mark Chen, Eric Sigler, Mateusz Litwin, Scott Gray, Benjamin Chess, Jack Clark, Christopher Berner, Sam McCandlish, Alec Radford, Ilya Sutskever, and Dario Amodei. 2020.
\newblock \href {https://proceedings.neurips.cc/paper_files/paper/2020/file/1457c0d6bfcb4967418bfb8ac142f64a-Paper.pdf} {Language models are few-shot learners}.
\newblock In \emph{Advances in Neural Information Processing Systems}, volume~33, pages 1877--1901. Curran Associates, Inc.

\bibitem[{Charniak and Johnson(2001)}]{charniak-johnson-2001-edit}
Eugene Charniak and Mark Johnson. 2001.
\newblock \href {https://aclanthology.org/N01-1016} {Edit detection and parsing for transcribed speech}.
\newblock In \emph{Second Meeting of the North {A}merican Chapter of the Association for Computational Linguistics}.

\bibitem[{Chen et~al.(2023)Chen, Papangelis, Tao, Kim, Rosenbaum, Liu, Yu, and Hakkani-Tur}]{chen-etal-2023-places}
Maximillian Chen, Alexandros Papangelis, Chenyang Tao, Seokhwan Kim, Andy Rosenbaum, Yang Liu, Zhou Yu, and Dilek Hakkani-Tur. 2023.
\newblock \href {https://doi.org/10.18653/v1/2023.findings-eacl.63} {{PLACES}: Prompting language models for social conversation synthesis}.
\newblock In \emph{Findings of the Association for Computational Linguistics: EACL 2023}, pages 844--868, Dubrovnik, Croatia. Association for Computational Linguistics.

\bibitem[{Colman and Healey(2011)}]{Colman2011TheDO}
Marcus Colman and Patrick G.~T. Healey. 2011.
\newblock \href {https://api.semanticscholar.org/CorpusID:11291221} {The distribution of repair in dialogue}.
\newblock \emph{Cognitive Science}, 33.

\bibitem[{Emrani and Hooshmand(2019)}]{Emrani2019}
Farnoosh Emrani and Mozhgan Hooshmand. 2019.
\newblock A conversation analysis of self-initiated self-repair structures in advanced iranian efl learners.
\newblock \emph{International Journal of Language Studies}, 13:57--76.

\bibitem[{Eric et~al.(2017)Eric, Krishnan, Charette, and Manning}]{eric-etal-2017-key}
Mihail Eric, Lakshmi Krishnan, Francois Charette, and Christopher~D. Manning. 2017.
\newblock \href {https://doi.org/10.18653/v1/W17-5506} {Key-value retrieval networks for task-oriented dialogue}.
\newblock In \emph{Proceedings of the 18th Annual {SIG}dial Meeting on Discourse and Dialogue}, pages 37--49, Saarbr{\"u}cken, Germany. Association for Computational Linguistics.

\bibitem[{Ferreira and Bailey(2004)}]{FERREIRA2004231}
Fernanda Ferreira and Karl~G.D. Bailey. 2004.
\newblock \href {https://doi.org/https://doi.org/10.1016/j.tics.2004.03.011} {Disfluencies and human language comprehension}.
\newblock \emph{Trends in Cognitive Sciences}, 8(5):231--237.

\bibitem[{Fox~Tree(2001)}]{foxtree2001}
Jean Fox~Tree. 2001.
\newblock \href {https://doi.org/10.3758/BF03194926} {Listeners' uses of um and uh in speech comprehension}.
\newblock \emph{Memory \& cognition}, 29:320--6.

\bibitem[{Goel et~al.(2023)Goel, Ammar, Gupta, Vashishtha, Sano, Surani, Chang, Choe, Greene, He, Nitisaroj, Trukhina, Paul, Shah, Shah, and Yu}]{goel-etal-2023-presto}
Rahul Goel, Waleed Ammar, Aditya Gupta, Siddharth Vashishtha, Motoki Sano, Faiz Surani, Max Chang, HyunJeong Choe, David Greene, Chuan He, Rattima Nitisaroj, Anna Trukhina, Shachi Paul, Pararth Shah, Rushin Shah, and Zhou Yu. 2023.
\newblock \href {https://doi.org/10.18653/v1/2023.emnlp-main.667} {{PRESTO}: A multilingual dataset for parsing realistic task-oriented dialogs}.
\newblock In \emph{Proceedings of the 2023 Conference on Empirical Methods in Natural Language Processing}, pages 10820--10833, Singapore. Association for Computational Linguistics.

\bibitem[{Hough and Schlangen(2015)}]{hough15_interspeech}
Julian Hough and David Schlangen. 2015.
\newblock \href {https://doi.org/10.21437/Interspeech.2015-264} {Recurrent neural networks for incremental disfluency detection}.
\newblock In \emph{Interspeech 2015}, pages 849--853.

\bibitem[{Li et~al.(2016)Li, Galley, Brockett, Gao, and Dolan}]{li-etal-2016-diversity}
Jiwei Li, Michel Galley, Chris Brockett, Jianfeng Gao, and Bill Dolan. 2016.
\newblock \href {https://doi.org/10.18653/v1/N16-1014} {A diversity-promoting objective function for neural conversation models}.
\newblock In \emph{Proceedings of the 2016 Conference of the North {A}merican Chapter of the Association for Computational Linguistics: Human Language Technologies}, pages 110--119, San Diego, California. Association for Computational Linguistics.

\bibitem[{Liu et~al.(2024)Liu, Fang, Vandyke, and Collier}]{liu-etal-2024-toad}
Yinhong Liu, Yimai Fang, David Vandyke, and Nigel Collier. 2024.
\newblock \href {https://doi.org/10.18653/v1/2024.findings-acl.494} {{TOAD}: Task-oriented automatic dialogs with diverse response styles}.
\newblock In \emph{Findings of the Association for Computational Linguistics: ACL 2024}, pages 8341--8356, Bangkok, Thailand. Association for Computational Linguistics.

\bibitem[{Marie(2023)}]{marie-2023-disfluency}
Benjamin Marie. 2023.
\newblock \href {https://doi.org/10.18653/v1/2023.findings-acl.728} {Disfluency generation for more robust dialogue systems}.
\newblock In \emph{Findings of the Association for Computational Linguistics: ACL 2023}, pages 11479--11488, Toronto, Canada. Association for Computational Linguistics.

\bibitem[{MetaAI(2024)}]{meta2024llama3}
MetaAI. 2024.
\newblock Llama 3.1: Instruction-tuned large language models.
\newblock \url{https://ai.meta.com/blog/meta-llama-3-1/}.
\newblock Accessed: 2024-07-23.

\bibitem[{OpenAI(2024)}]{openai2024gpt4o}
OpenAI. 2024.
\newblock Gpt-4.o: Optimized language models for multimodal applications.
\newblock \url{https://openai.com/index/hello-gpt-4o/}.
\newblock Accessed: 2024-05-10.

\bibitem[{Passali et~al.(2022)Passali, Mavropoulos, Tsoumakas, Meditskos, and Vrochidis}]{passali-etal-2022-lard}
Tatiana Passali, Thanassis Mavropoulos, Grigorios Tsoumakas, Georgios Meditskos, and Stefanos Vrochidis. 2022.
\newblock \href {https://aclanthology.org/2022.lrec-1.249} {{LARD}: Large-scale artificial disfluency generation}.
\newblock In \emph{Proceedings of the Thirteenth Language Resources and Evaluation Conference}, pages 2327--2336, Marseille, France. European Language Resources Association.

\bibitem[{Paszke et~al.(2019)Paszke, Gross, Massa, Lerer, Bradbury, Chanan, Killeen, Lin, Gimelshein, Antiga, Desmaison, Kopf, Yang, DeVito, Raison, Tejani, Chilamkurthy, Steiner, Fang, Bai, and Chintala}]{NEURIPS2019_bdbca288}
Adam Paszke, Sam Gross, Francisco Massa, Adam Lerer, James Bradbury, Gregory Chanan, Trevor Killeen, Zeming Lin, Natalia Gimelshein, Luca Antiga, Alban Desmaison, Andreas Kopf, Edward Yang, Zachary DeVito, Martin Raison, Alykhan Tejani, Sasank Chilamkurthy, Benoit Steiner, Lu~Fang, Junjie Bai, and Soumith Chintala. 2019.
\newblock \href {https://proceedings.neurips.cc/paper_files/paper/2019/file/bdbca288fee7f92f2bfa9f7012727740-Paper.pdf} {Pytorch: An imperative style, high-performance deep learning library}.
\newblock In \emph{Advances in Neural Information Processing Systems}, volume~32. Curran Associates, Inc.

\bibitem[{Plejert(2004)}]{Plejert2004FixWhatsNotBroken}
Charlotta Plejert. 2004.
\newblock \emph{To Fix What’s Not Broken: Repair Strategies in Non-Native and Native English Conversation}.
\newblock Ph.D. thesis.

\bibitem[{Raffel et~al.(2020)Raffel, Shazeer, Roberts, Lee, Narang, Matena, Zhou, Li, and Liu}]{raffel2019exploring}
Colin Raffel, Noam Shazeer, Adam Roberts, Katherine Lee, Sharan Narang, Michael Matena, Yanqi Zhou, Wei Li, and Peter~J Liu. 2020.
\newblock \href {https://doi.org/10.48550/arXiv.1910.10683} {Exploring the limits of transfer learning with a unified text-to-text transformer}.
\newblock \emph{Journal of Machine Learning Research}, 21(140):1--67.

\bibitem[{Rastogi et~al.(2020)Rastogi, Zang, Sunkara, Gupta, and Khaitan}]{rastogi2020towards}
Abhinav Rastogi, Xiaoxue Zang, Srinivas Sunkara, Raghav Gupta, and Pranav Khaitan. 2020.
\newblock \href {https://doi.org/10.1609/aaai.v34i05.6394} {Towards scalable multi-domain conversational agents: The schema-guided dialogue dataset}.
\newblock In \emph{Proceedings of the AAAI Conference on Artificial Intelligence}, volume~34, pages 8689--8696.

\bibitem[{Seyfeddinipur(2006)}]{Seyfeddinipur2006Disfluency}
M.~Seyfeddinipur. 2006.
\newblock \href {https://doi.org/10.17617/2.59337} {\emph{Disfluency: Interrupting Speech and Gesture}}.
\newblock Ph.D. thesis, Radboud University Nijmegen, Nijmegen.

\bibitem[{Shah et~al.(2018)Shah, Hakkani-T{\"u}r, Liu, and T{\"u}r}]{shah-etal-2018-bootstrapping}
Pararth Shah, Dilek Hakkani-T{\"u}r, Bing Liu, and Gokhan T{\"u}r. 2018.
\newblock \href {https://doi.org/10.18653/v1/N18-3006} {Bootstrapping a neural conversational agent with dialogue self-play, crowdsourcing and on-line reinforcement learning}.
\newblock In \emph{Proceedings of the 2018 Conference of the North {A}merican Chapter of the Association for Computational Linguistics: Human Language Technologies, Volume 3 (Industry Papers)}, pages 41--51, New Orleans - Louisiana. Association for Computational Linguistics.

\bibitem[{Shriberg(1994)}]{Shriberg1994PreliminariesTA}
Elizabeth Shriberg. 1994.
\newblock \href {https://api.semanticscholar.org/CorpusID:62654138} {Preliminaries to a theory of speech disfluencies}.

\bibitem[{Sparks(1995)}]{Sparks1995StructureOS}
R.~Sparks. 1995.
\newblock \emph{The Structure of Self-Repair in English Conversation}.
\newblock Ph.D. thesis, Department of Linguistics, University of Colorado, Boulder, CO.

\bibitem[{Stacey et~al.(2024)Stacey, Cheng, Torr, Guigue, Driesen, Coca, Gaynor, and Johannsen}]{stacey-etal-2024-lucid}
Joe Stacey, Jianpeng Cheng, John Torr, Tristan Guigue, Joris Driesen, Alexandru Coca, Mark Gaynor, and Anders Johannsen. 2024.
\newblock \href {https://doi.org/10.18653/v1/2024.naacl-srw.8} {{LUCID}: {LLM}-generated utterances for complex and interesting dialogues}.
\newblock In \emph{Proceedings of the 2024 Conference of the North American Chapter of the Association for Computational Linguistics: Human Language Technologies (Volume 4: Student Research Workshop)}, pages 56--74, Mexico City, Mexico. Association for Computational Linguistics.

\bibitem[{Wolf et~al.(2020)Wolf, Debut, Sanh, Chaumond, Delangue, Moi, Cistac, Rault, Louf, Funtowicz, Davison, Shleifer, von Platen, Ma, Jernite, Plu, Xu, Le~Scao, Gugger, Drame, Lhoest, and Rush}]{wolf-etal-2020-transformers}
Thomas Wolf, Lysandre Debut, Victor Sanh, Julien Chaumond, Clement Delangue, Anthony Moi, Pierric Cistac, Tim Rault, Remi Louf, Morgan Funtowicz, Joe Davison, Sam Shleifer, Patrick von Platen, Clara Ma, Yacine Jernite, Julien Plu, Canwen Xu, Teven Le~Scao, Sylvain Gugger, Mariama Drame, Quentin Lhoest, and Alexander Rush. 2020.
\newblock \href {https://doi.org/10.18653/v1/2020.emnlp-demos.6} {Transformers: State-of-the-art natural language processing}.
\newblock In \emph{Proceedings of the 2020 Conference on Empirical Methods in Natural Language Processing: System Demonstrations}, pages 38--45, Online. Association for Computational Linguistics.

\bibitem[{Yang et~al.(2020)Yang, Yang, and Ma}]{yang-etal-2020-planning}
Jingfeng Yang, Diyi Yang, and Zhaoran Ma. 2020.
\newblock \href {https://doi.org/10.18653/v1/2020.emnlp-main.113} {Planning and generating natural and diverse disfluent texts as augmentation for disfluency detection}.
\newblock In \emph{Proceedings of the 2020 Conference on Empirical Methods in Natural Language Processing (EMNLP)}, pages 1450--1460, Online. Association for Computational Linguistics.

\bibitem[{Zang et~al.(2020)Zang, Rastogi, Sunkara, Gupta, Zhang, and Chen}]{zang-etal-2020-multiwoz}
Xiaoxue Zang, Abhinav Rastogi, Srinivas Sunkara, Raghav Gupta, Jianguo Zhang, and Jindong Chen. 2020.
\newblock \href {https://doi.org/10.18653/v1/2020.nlp4convai-1.13} {{M}ulti{WOZ} 2.2 : A dialogue dataset with additional annotation corrections and state tracking baselines}.
\newblock In \emph{Proceedings of the 2nd Workshop on Natural Language Processing for Conversational AI}, pages 109--117, Online. Association for Computational Linguistics.

\bibitem[{Zayats et~al.(2016)Zayats, Ostendorf, and Hajishirzi}]{zayats16_interspeech}
Vicky Zayats, Mari Ostendorf, and Hannaneh Hajishirzi. 2016.
\newblock \href {https://doi.org/10.21437/Interspeech.2016-1247} {Disfluency detection using a bidirectional lstm}.
\newblock In \emph{Interspeech 2016}, pages 2523--2527.

\bibitem[{Zayats et~al.(2019)Zayats, Tran, Wright, Mansfield, and Ostendorf}]{zayats19_interspeech}
Vicky Zayats, Trang Tran, Richard Wright, Courtney Mansfield, and Mari Ostendorf. 2019.
\newblock \href {https://doi.org/10.21437/Interspeech.2019-3134} {Disfluencies and human speech transcription errors}.
\newblock In \emph{Interspeech 2019}, pages 3088--3092.

\bibitem[{Zhang et~al.(2020)Zhang, Sun, Galley, Chen, Brockett, Gao, Gao, Liu, and Dolan}]{zhang-etal-2020-dialogpt}
Yizhe Zhang, Siqi Sun, Michel Galley, Yen-Chun Chen, Chris Brockett, Xiang Gao, Jianfeng Gao, Jingjing Liu, and Bill Dolan. 2020.
\newblock \href {https://doi.org/10.18653/v1/2020.acl-demos.30} {{DIALOGPT} : Large-scale generative pre-training for conversational response generation}.
\newblock In \emph{Proceedings of the 58th Annual Meeting of the Association for Computational Linguistics: System Demonstrations}, pages 270--278, Online. Association for Computational Linguistics.

\end{thebibliography}

\appendix
\section{Appendix}

\subsection{Human Evaluation Setup}
\label{sec:appendix-human-eval}

We conducted human evaluation to assess the quality of dialogs across various dimensions in both the DiscoDrive Dataset and the KVRET Dataset \citep{eric-etal-2017-key}. The evaluation consisted of both intrinsic and comparative assessments using the following metrics:

\subsubsection{Intrinsic Evaluation Metrics}
Annotators rated each dialogue individually using a 5-point Likert scale across the following metrics:

\begin{itemize}
    \item \textbf{Dialogue Naturalness (Conversational Realism):} Measures how natural and human-like the conversation feels. 
    \\ \textit{Task:} Rate from 1 (robotic/artificial) to 5 (completely natural).

    \item \textbf{Dialogue Coherence (Logical Flow):} Assesses the logical progression of the conversation and whether each response follows logically from the previous one.
    \\ \textit{Task:} Rate from 1 (disjointed/confusing) to 5 (clear/logical flow).

    \item \textbf{Engagement and Dynamism (Conversational Liveliness):} Evaluates how engaging and dynamic the conversation feels, capturing attention or feeling flat.
    \\ \textit{Task:} Rate from 1 (boring/monotonous) to 5 (lively/engaging).

    \item \textbf{Dialogue Consistency:} Measures the consistency of responses, ensuring that there are no contradictions in the conversation.
    \\ \textit{Task:} Rate from 1 (frequent contradictions) to 5 (fully consistent).

    \item \textbf{On-topic Relevance:} Assesses how well the conversation stays on topic without introducing irrelevant information.
    \\ \textit{Task:} Rate from 1 (off-topic frequently) to 5 (fully on-topic).

    \item \textbf{Disfluency Realism (Naturalness of Driver's Disfluencies):} If present, rates the naturalness of the disfluencies in the driver’s utterances.
    \\ \textit{Task:} Rate from 1 (forced/unnecessary disfluencies) to 5 (natural disfluencies that enhance realism).
\end{itemize}

\subsubsection{Comparative Evaluation Metrics}
In the comparative evaluation, annotators compared pairs of dialogues (one from the DiscoDrive dataset and one from the KVRET dataset \citep{eric-etal-2017-key}). Each annotator was asked to choose which dialogue was better across the following categories:

\begin{itemize}
    \item \textbf{Overall Dialogue Quality Preference:} Which dialogue feels more natural and human-like overall?
    \\ \textit{Task:} Choose between the DiscoDrive Dataset or the KVRET Dataset \citep{eric-etal-2017-key}.

    \item \textbf{Conversational Naturalness Preference:} Which dialogue flows more like a natural conversation?
    \\ \textit{Task:} Choose between the DiscoDrive Dataset or the KVRET Dataset. \citep{eric-etal-2017-key}.

    \item \textbf{Task Effectiveness Preference:} Which dialogue better helps the driver achieve their goal?
    \\ \textit{Task:} Choose between the DiscoDrive Dataset or the KVRET Dataset. \citep{eric-etal-2017-key}.

    \item \textbf{Human-Likeness of Driver's Utterance Preference:} Which driver's utterances feel more realistic and human-like?
    \\ \textit{Task:} Choose between the DiscoDrive Dataset or the KVRET Dataset. \citep{eric-etal-2017-key}.

    \item \textbf{Engagement Preference:} Which conversation is more lively and engaging?
    \\ \textit{Task:} Choose between the DiscoDrive Dataset or the KVRET Dataset. \citep{eric-etal-2017-key}.
\end{itemize}

\subsubsection{Evaluation Guidelines for Dynamic vs. Post-Hoc Disfluency Integration}
Annotators rated KVRET dialogs modified by both the LARD method \citep{passali-etal-2022-lard} (post-hoc insertion) and dialogs from DiscoDrive dataset on a 5-point Likert scale based on the following criteria:

\begin{itemize} \item \textbf{Naturalness:} Assesses how naturally the disfluencies are integrated into the dialog.
\ \textit{Task:} Rate from 1 (disfluencies appear forced or out-of-place) to 5 (disfluencies blend seamlessly with natural speech patterns).

\item \textbf{Appropriateness:} Evaluates whether the inserted disfluencies are contextually relevant and enhance the conversational flow rather than disrupt it.  
\\ \textit{Task:} Rate from 1 (disfluencies are contextually irrelevant or inappropriate) to 5 (disfluencies are perfectly suited to the dialog context).
\item \textbf{Clarity:} Measures whether the addition of disfluencies impairs the overall clarity and comprehensibility of the conversation.  
\\ \textit{Task:} Rate from 1 (disfluencies significantly hinder understanding) to 5 (disfluencies do not detract from clarity).
\end{itemize}

\subsubsection{Evaluation Procedure}
Two human evaluators independently rated 140 dialogs from each dataset. For the KVRET Dataset \citep{eric-etal-2017-key}, 100 dialogs were sampled from the training set, and 20 each from the validation and test sets. For the DiscoDrive Dataset, 20 dialogs were selected from each of the seven domains, with 4 dialogs representing turn lengths of 6, 8, 10, 12, and 14 turns. Stratified sampling ensured diversity in both content and length. For comparison with post-hoc disfluency generation methods, we used the same dialogs with LARD \citep{passali-etal-2022-lard} method to incorporate disfluencies in the KVRET \citep{eric-etal-2017-key} dialogs.

\subsection{Examples of Post-Hoc Disfluency Insertion using LARD method \citep{passali-etal-2022-lard}}
\label{sec:post-hoc}

The LARD (Large-scale Artificial Disfluency Generation) method \citep{passali-etal-2022-lard} generates synthetic disfluencies from fluent text using a post-hoc approach. This method operates by modifying fluent sequences according to three primary operations: repetitions, replacements, and restarts. In each case, the process involves a degree of random selection to determine which tokens to alter, without considering the full conversational context. 

\subsubsection*{Repetition Generation} For repetitions, the method randomly selects an index within a fluent sequence and repeats one or more tokens. For instance, given the fluent sentence: \begin{quote} \texttt{will it be raining in the next 7 days.} \end{quote} The disfluent version produced is: \begin{quote} \texttt{will it be raining in the next [7 days + 7 days].} \end{quote} Here, the phrase "7 days" is repeated to simulate a natural hesitation.

\subsubsection*{Replacement Generation} In the replacement operation, a candidate token (e.g., a noun, verb, or adjective) is randomly selected from the fluent sequence. A synonym or antonym is then chosen as a substitute, optionally with the insertion of a repair cue. For example, given: \begin{quote} \texttt{show me the closest location where i can get chinese food.} \end{quote} The disfluent version generated is: \begin{quote} \texttt{show me the closest location where [{no sorry} + the nearest restaurant] where i can get chinese food.} \end{quote} 

\subsubsection*{Restart Generation} For restarts, the method assumes the availability of two distinct fluent sequences. It randomly splits the first sequence at a selected point and then replaces the remainder with the entirety of the second sequence. For example, given: \begin{quote} Sequence 1: \texttt{Set a reminder that I have a lab appointment with my aunt next Wednesday at 1pm.} \\ Sequence 2: \texttt{Check to see if it will be windy in brentwood the next few days.} \end{quote} The disfluent version generated is: \begin{quote} \texttt{[Set a reminder that I + ] Check to see if it will be windy in brentwood the next few days.} \end{quote} In this example, the first sequence is abruptly terminated, and the second sequence is inserted, reflecting a complete restart of the utterance.

\subsection{Prompt Design Examples} \label{sec:appendix-prompts}

This section provides the detailed prompts used for generating realistic driver-AI dialogs. These prompts were designed to guide the generative model in producing naturalistic conversations, with a focus on disfluency integration for the Driver role and concise, task-specific responses for the Car AI. 

\subsubsection{Driver Prompt}
The Driver prompts were designed to encourage the generation of realistic, disfluent speech patterns as shown in \ref{tab:driver1} and \ref{tab:driver2}:

\begin{table*}[ht!]
\centering
\begin{minipage}{\textwidth}
\noindent\fbox{
\begin{minipage}{\textwidth}
\textbf{Driver Prompt (for follow-up question generation):}  

You are a human driver interacting with a car AI system. Ask a short, relevant follow-up question based on the AI’s response. Keep the question concise and avoid repetition or unnecessary details.

Incorporate one disfluency into your question to make the conversation more lifelike. Occasionally, you may use none or even two or more disfluencies if it fits the flow naturally. Randomly select from the different types of disfluencies to keep the conversation dynamic. Make sure each disfluency fits naturally and enhances the authenticity of the question without making it too long or incomprehensible.

\textbf{Types of Disfluencies to Use} (Refer to the examples to understand how to effectively incorporate them):

\begin{itemize}
    \item \textbf{a) Repetitions:} Briefly repeat a word or phrase to emphasize a point or show hesitation. \\
    Example: ``I feel like, I feel like we’re going in the wrong direction.''
    
    \item \textbf{b) False Starts:} Begin a sentence, then adjust or rethink your wording, reflecting a natural thought process. \\
    Example: ``I was planning to—actually, wait, do we need gas first?''
    
    \item \textbf{c) Pauses:} Insert brief pauses using ``...'' to reflect hesitation or thoughtfulness. \\
    Example: ``So, we’re going to... um, the restaurant?''
    
    \item \textbf{d) Corrections:} Correct yourself when realizing a mistake or clarifying a detail. \\
    Example: ``I’ll pick you up at 6—oh, no, sorry, 6:30.''
    
    \item \textbf{e) Filler Words:} Use casual words like ``um,'' ``uh,'' or ``you know'' to fill pauses and soften the delivery. \\
    Example: ``Can you, um, tell me how far we are from the destination?''

\end{itemize}
\end{minipage}
}
\end{minipage}
\caption{Driver Prompt for follow-up question generation}
\label{tab:driver1}
\end{table*}

\begin{table*}[ht!]
\centering
\begin{minipage}{\textwidth}
\noindent\fbox{
\begin{minipage}{\textwidth}
\textbf{Driver Prompt (for concluding the conversation at the last turn:}  

You are a human driver interacting with a car AI system. Your task is to casually wrap up the conversation in a brief, natural, and short manner without formalities.

Keep the conclusion concise, ideally with one disfluency. However, feel free to use none or multiple disfluencies if it fits the flow better. Randomly select from different types of disfluencies to maintain a dynamic and engaging wrap-up. Make sure disfluencies fit naturally with the context and don’t make the conclusion confusing or too long.

Use the following types of disfluencies (refer to the examples to see how they can be used naturally):

\begin{itemize}
    \item \textbf{a) Repetitions:} Briefly repeat a word or phrase to emphasize an idea or express hesitation. \\
    Example: ``I feel like, I feel like we’re going the wrong way.''
    
    \item \textbf{b) False Starts:} Begin a sentence but change your mind midway. \\
    Example: ``I was going to say—wait, can we take a different route?''
    
    \item \textbf{c) Pauses:} Use ``...'' to show a brief pause or hesitation. \\
    Example: ``I think we’ll be there... um, soon.''
    
    \item \textbf{d) Corrections:} Correct yourself when you realize a mistake or need to clarify. \\
    Example: ``We’re meeting at 5—oh, no, 6.''
    
    \item \textbf{e) Filler Words:} Use casual words to fill pauses and buy thinking time. \\
    Example: ``So, like, what time is it?''
\end{itemize}
\end{minipage}
}
\end{minipage}
\caption{Driver Prompt for concluding the conversation at the last turn}
\label{tab:driver2}
\end{table*}

\subsubsection{Car AI Prompt}
The Car AI prompts ensured concise, task-specific responses as shown in \ref{tab:ai1} and \ref{tab:ai2}:

\begin{table*}[t]
\centering
\begin{minipage}{\textwidth}
\noindent\fbox{
\begin{minipage}{\textwidth}
\textbf{Car AI Prompt (for responding to the queries from driver):}

You are a car AI system. Your role is to assist the driver by providing accurate, clear, and task-specific information. Keep your responses short, friendly, and directly focused on the driver’s request. Avoid unnecessary details while ensuring your tone remains approachable and helpful. Your goal is to offer practical assistance in a concise and efficient manner, enhancing the driver’s experience without distraction.
\end{minipage}
}
\end{minipage}
\caption{Car AI Prompt for responding to the queries from the driver}
\label{tab:ai1}
\end{table*}

\begin{table*}[t]
\centering
\begin{minipage}{\textwidth}
\noindent\fbox{
\begin{minipage}{\textwidth}
\textbf{Car AI Prompt (for concluding the conversation at the last turn):}  

You are a car AI system. When the driver concludes the conversation, respond with a brief, friendly acknowledgment. You may include a short, polite closing remark or summary to wrap up the interaction. Keep it concise and avoid asking further questions or offering unnecessary details.
\end{minipage}
}
\end{minipage}
\caption{Car AI Prompt for concluding the conversation at the last turn}
\label{tab:ai2}
\end{table*}

\subsection{Example Dialogs from DiscoDrive Dataset}

The following example dialogs from our DiscoDrive dataset demonstrate the interaction between a driver and the car AI, showcasing the natural use of disfluencies in the driver's utterances. 

\begin{table*}[ht!]
\centering
\resizebox{\textwidth}{!}{%
\begin{tabular}{p{3cm} p{12cm}}  
\hline
\textbf{Speaker} & \textbf{Utterance} \\
\hline
Driver & "Um, can you... show me a list of nearby charging stations, please?" \\
Car AI & "Help is on the way. I'm checking the nearest charging stations. Here are a few options within a 5-mile radius: \newline
1. 'QuickCharge' - 2.5 miles away, 10 minutes \newline
2. 'Electric Oasis' - 3.2 miles away, 12 minutes \newline
3. 'GreenHub' - 4.5 miles away, 18 minutes \newline
Which one would you like me to navigate to?" \\
Driver & "I, uh, need to know more about QuickCharge... how is the, uh, condition of the charging stations?" \\
Car AI & "I've checked the status of QuickCharge. Here's what I found: \newline
The QuickCharge station has 6 available charging ports, and 4 are currently in use. The station is equipped with a 100kW fast charger. The station's condition is rated as 'Good' with a cleanliness rating of 8/10. Would you like to navigate to QuickCharge now?" \\
Driver & "Wait, so... four of the charging ports are occupied, right?" \\
Car AI & "Yes, that's correct. Four of the six charging ports at QuickCharge are currently in use." \\
Driver & "Yeah, four occupied... yeah, that might not be ideal. So, like, we'll just stick with the other options, I guess." \\
Car AI & "I'll navigate to Electric Oasis, which is 3.2 miles away and should only take around 12 minutes to reach." \\
\hline
\end{tabular}%
}
\caption{Safety and emergency assistance - Finding nearby charging stations.}
\label{tab:charging_stations_dialog}
\end{table*}

\begin{table*}[ht!]
\centering
\begin{tabular}{p{3cm} p{13cm}} 
\hline
\textbf{Speaker} & \textbf{Utterance} \\
\hline
Driver & "Uh, how often do I need to get an oil change, considering my car's current mileage?" \\
Car AI & "Your car's recommended oil change interval is 5,000 miles. Based on your current mileage, I'll set a reminder for an oil change at 5,500 miles. Would you like me to send a notification to your phone?" \\
Driver & "Wait, that's... um, a little soon, right? Can I get the reminder set for like, a week before the oil change, not exactly at 5,500 miles?" \\
Car AI & "I'll set the reminder for a week before the oil change, which will be at approximately 5,000 miles. That way, you'll have enough time to schedule the oil change before reaching the recommended interval. Would you like me to send a notification to your phone?" \\
Driver & "Yeah, that sounds good... I mean, yeah, send the notification to my phone. Thanks for checking that for me." \\
Car AI & "You're welcome. I've set a reminder for an oil change at approximately 5,000 miles, with a notification sent to your phone a week prior. I'll keep an eye on your mileage to ensure you stay on schedule." \\
\hline
\end{tabular}
\caption{Car maintenance and diagnostics : Discussion on Oil Change Reminder.}
\end{table*}

\begin{table*}[ht!]
\centering
\begin{tabular}{p{3cm} p{13cm}} 
\hline
\textbf{Speaker} & \textbf{Utterance} \\
\hline
Driver & "Hey, can you, uh, suggest any popular wellness centers near our current location?" \\
Car AI & "Welcome to Kerala. I've found a few popular wellness centers near your current location. Would you like me to suggest 'Kerala Ayurveda Limited' in Kovalam or 'Somatheeram Ayurveda Village' in Thiruvananthapuram? I can provide directions and more information if you're interested." \\
Driver & "What's, uh, the difference between Kerala Ayurveda Limited and Somatheeram Ayurveda Village, exactly?" \\
Car AI & "Kerala Ayurveda Limited focuses on traditional Ayurvedic treatments and products. It's a more modern take on the ancient practices. Somatheeram Ayurveda Village, on the other hand, is a holistic wellness resort offering Ayurvedic treatments, yoga, and a peaceful atmosphere amidst nature. It's more of a retreat experience. Would you like me to provide directions to either of these locations or help you decide based on your preferences?" \\
Driver & "So, um, you're saying Somatheeram Ayurveda Village is more of a retreat, right?" \\
Car AI & "That's correct. Somatheeram Ayurveda Village is a holistic wellness resort offering a peaceful retreat experience with Ayurvedic treatments, yoga, and a serene atmosphere amidst nature. If you're interested in a relaxing getaway, it might be the perfect choice. Would you like me to provide directions to Somatheeram Ayurveda Village or help you compare it with Kerala Ayurveda Limited further?" \\
Driver & "So, like... I think we've got a good idea of our options now." \\
Car AI & "Thank you for chatting with me about wellness centers in Kerala. I'm glad I could help you understand the options. I'll keep assisting you until you're ready to explore the area." \\
\hline
\end{tabular}
\caption{Local and on-route attractions and activities - Discussion about wellness centers in Kerala.}
\end{table*}

\begin{table*}[ht!]
\centering
\begin{tabular}{p{3cm} p{13cm}} 
\hline
\textbf{Speaker} & \textbf{Utterance} \\
\hline
Driver & "Um, how's the, uh, weather forecast looking for our route from Surat to Daman, specifically around the coastal areas?" \\
Car AI & "For your route from Surat to Daman, I've checked the current weather conditions. There's a moderate chance of light to moderate rain showers along the coastal areas, particularly near Daman. I recommend keeping an eye on the road conditions and slowing down if the rain becomes heavy." \\
Driver & "You know, that's helpful, but... what's the expected timing of these rain showers, exactly?" \\
Car AI & "For your route from Surat to Daman, the rain showers are expected to start around 2 pm and continue until 6 pm, with a brief break in between." \\
Driver & "What's, uh, the chance of heavy rain during those, you know, peak hours from 2 to 6 pm?" \\
Car AI & "I've checked the latest forecast. There's a 30\% chance of heavy rain during the peak hours from 2 pm to 6 pm along the coastal areas near Daman." \\
Driver & "Yeah, good to know... so we should be prepared for some rain, huh?" \\
Car AI & "That's a good idea to be prepared. We'll keep an eye on the road conditions and slow down if needed. Our route will likely be a bit slippery during the rain, so let's stay safe and cautious." \\
\hline
\end{tabular}
\caption{Weather - Weather forecast for a trip from Surat to Daman.}
\end{table*}

\begin{table*}[ht!]
\centering
\begin{tabular}{p{3cm} p{13cm}} 
\hline
\textbf{Speaker} & \textbf{Utterance} \\
\hline
Driver & "Uh, do you have, um, any audiobooks on jazz history that I could listen to on our way to the concert?" \\
Car AI & "We have a great selection of audiobooks on jazz history. Let me suggest a few popular ones: The Jazz Standards: A Guide to the Repertoire by Al Gioia, Jazz: A Century of Change by John Edward Hasse, and The Musician's Guide to Reading and Writing Music by Gary E. Anderson. Which one would you like to listen to?" \\
Driver & "So, The Jazz Standards sounds... you know, interesting, but is Jazz: A Century of Change more focused on the, uh, historical aspects of jazz?" \\
Car AI & "Jazz: A Century of Change by John Edward Hasse indeed focuses on the evolution of jazz throughout its history, covering key events, musicians, and styles that shaped the genre. Would you like to listen to a preview or start the audiobook?" \\
Driver & "Wait, so it covers key events and musicians... you know, that would be really helpful for the concert tonight. Can I, um, get a list of the musicians it mentions?" \\
Car AI & "Key musicians covered in the audiobook include Louis Armstrong, Duke Ellington, Charlie Parker, Miles Davis, John Coltrane, and Billie Holiday." \\
Driver & "That's, that's really helpful to know. I think we'll go with the preview of Jazz: A Century of Change... yeah, sounds good." \\
Car AI & "I've loaded the preview of Jazz: A Century of Change by John Edward Hasse for you to listen to on our way to the concert. Enjoy the audiobook and have a great time at the concert tonight." \\
\hline
\end{tabular}
\caption{Entertainment - Discussion on Jazz History Audiobooks.}
\end{table*}

\begin{table*}[ht!]
\centering
\begin{tabular}{p{3cm} p{13cm}} 
\hline
\textbf{Speaker} & \textbf{Utterance} \\
\hline
Driver & "You know, can you, um, show me how to turn on the cabin air filtration system, please?" \\
Car AI & "Press the 'Settings' button on your steering wheel. Then, select 'Comfort' and choose 'Air Quality' from the menu. Finally, toggle the 'Cabin Air Filtration' switch to on. You can also set the system to automatically turn on when the engine starts." \\
Driver & "Okay, so I press the 'Settings' button, select 'Comfort', and then... 'Air Quality' – that's all pretty straightforward, but, uh, how do I know the system is actually working once it's on?" \\
Car AI & "To check if the cabin air filtration system is working, look for the air quality indicator on your dashboard. It should display a green or blue icon when the system is activated and the air quality is good. You can also check the system's status in the 'Air Quality' menu under 'Comfort' settings." \\
Driver & "Alright, I think I got it... yeah, I'm gonna go check the dashboard now." \\
Car AI & "The cabin air filtration system should now be active. You can check its status on the dashboard or in the 'Air Quality' menu. If you need anything else, feel free to ask." \\
\hline
\end{tabular}
\caption{Car functions - Instruction on Activating Cabin Air Filtration System.}
\end{table*}

\begin{table*}[ht!]
\centering
\resizebox{\textwidth}{!}{%
\begin{tabular}{p{3cm} p{12cm}}  
\hline
\textbf{Speaker} & \textbf{Utterance} \\
\hline
Driver & "So, you know, can you, um, show me the banks around me that are open on a Sunday?" \\
Car AI & "Affirmative, driver. I'm accessing nearby bank locations in Hyderabad. There are two banks open on a Sunday within a 5-kilometer radius from your current location. The SBI Main Branch on Abids Road is open from 10 am to 4 pm, and the ICICI Bank on Banjara Hills is open from 9 am to 5 pm. Shall I provide directions to either location?" \\
Driver & "I feel like, the SBI one is closer... can you, uh, give me the exact directions to the SBI Main Branch, please?" \\
Car AI & "Affirmative, driver. I'm providing you with turn-by-turn directions to the SBI Main Branch on Abids Road. \newline
Turn left at the upcoming signal onto MG Road. \newline
Continue on MG Road for 0.5 kilometers. \newline
Turn right onto Abids Road. \newline
The SBI Main Branch will be on your left in 300 meters. \newline
Please follow the directions, and you'll arrive at the SBI Main Branch. Estimated time of arrival is 5 minutes." \\
Driver & "Alright, got it... so, I think I'll just head to the SBI Main Branch now. Thanks for the help." \\
Car AI & "You're welcome, driver. Enjoy your visit to the SBI Main Branch. Safe travels." \\
\hline
\end{tabular}%
}
\caption{Navigation: Finding nearby banks open on a Sunday.}
\label{tab:banks_sunday_dialog}
\end{table*}

\end{document}